\documentclass[letterpaper, 10 pt, conference]{ieeeconf}  

\IEEEoverridecommandlockouts                              

\usepackage{float}
\usepackage[style=base]{caption}
\usepackage{subcaption}
\usepackage{graphics} 
\usepackage{epsfig} 
\usepackage[export]{adjustbox}
\usepackage[normalem]{ulem}
\usepackage{gensymb}

\usepackage{times} 
\usepackage{amsmath} 
\usepackage{amssymb}  
\usepackage{amsfonts}
\usepackage{amsbsy}
\usepackage{relsize}
\usepackage{color}
\usepackage{cite}
\usepackage{xspace}
\DeclareRobustCommand{\eg}{e.g.\@\xspace}
\DeclareRobustCommand{\ie}{i.e.\@\xspace}

\renewcommand{\P}[1]{P \big ( #1 \big )}
\newcommand{\W}[1]{W \big ( #1 \big )}
\newcommand{\E}[1]{\mathbb{E} \Big [  #1 \Big ]}
\newcommand{\Ls}[1]{L \big ( #1 \big )}
\newcommand{\risk}[1]{r \big ( #1 \big )}
\newcommand{\Symbol}[1]{\ensuremath{\mathcal{#1}}}

\newcommand{\argmax}{\mathop{\mathrm{argmax}}}
\newcommand{\argmin}{\mathop{\mathrm{min}}}
\newtheorem{definition}{Definition}{\bfseries}{\rmfamily}
\title{\LARGE \bf
Towards Real-Time Search Planning in Subsea Environments
}
\author{James McMahon$^{1}$, Harun Yetkin, Artur Wolek, Zachary J. Waters, Daniel J. Stilwell 
\thanks{*This work was supported by ONR Code 32}
\thanks{$^{1}$Corresponding author: james.mcmahon@nrl.navy.mil}%
\thanks{James McMahon and Zachary J. Waters are with the US Naval Research Laboratory, Code 7130, Washington DC, USA }%
\thanks{Artur Wolek is an ASEE Postdoctoral Fellow, Washington, D.C., USA }%
\thanks{Harun Yetkin and Daniel Stilwell are with the Bradley Department of Electrical and Computer Engineering, Virginia Tech, Blacksburg, VA, USA }%
}

\begin{document}

\maketitle
\thispagestyle{empty}
\pagestyle{empty}

\begin{abstract}

We address the challenge of computing search paths in real-time for subsea applications where the goal is to locate an unknown number of targets on the seafloor.  Our approach maximizes a formal definition of search effectiveness given finite search effort.   We account for false positive measurements and variation in the performance of the search sensor due to geographic variation of the seafloor.  We compare near-optimal search paths that can be computed in real-time with optimal search paths for which real-time computation is infeasible.  We show how  sonar data acquired  for locating targets at a specific location can also be used to characterize the performance of the search sonar at that location.  Our approach is illustrated with numerical experiments where search paths are planned using sonar data previously acquired from Boston Harbor.

\end{abstract}

\section{Introduction}

This paper addresses algorithms for real-time computation of  robotic search paths for subsea applications.  The goal is to find an unknown number of stationary targets distributed in a bounded area in finite time and/or finite distance.  Our contributions in this paper build upon prior work reported in \cite{yetkin.etal.oceans2015, yetkin.etal.oceans2016} in which we developed decision-theoretic cost functions that can be used to evaluate candidate search paths.  We assume herein, as we did in prior work, that the search sensor produces false positive measurements of targets that do not exist, and that the performance of the sensor is affected by spatial variation of the environment.  That is, in some locations, the search sensor performs better or worse than in other locations. 

We assume that the search sensor has the typical characteristics of a side-scan sonar, which is often used for imaging the seafloor.  Specifically, the system studied in this work performs \textit{synthetic aperture sonar processing} which requires the vehicle to travel in straight lines. This motion enables the vehicle to perform signal processing that greatly improves the quality of the collected data; perturbations in the straight-line motion of the vehicle significantly distort the resulting image ~\cite{chapple2008automated,Houston2008,Hayes2009}). Thus, a search path should be a set of straight lines and the number of turns should be small.  Performance of the search sensor at a location is dependent on the environment (e.g., seafloor characteristics) at the location, and our path planning approach accounts for a probabilistic  description of the environment.  The local environment, relative to sensor performance, can be inferred from sensor data.  Thus a search mission can be adjusted in real-time to account for information about the environment that is acquired during a search mission.  We present preliminary results on estimating sensor performance, and we illustrate our approach using a real data set abstracted from measurements made by an autonomous underwater vehicle (AUV) in the Boston Harbor. While the approach presented in this work is applied to an AUV that performs synthetic aperture sonar processing, the work may also be relevant to other unmanned systems (e.g. aircraft that  perform synthetic aperture radar processing~\cite{ouchi2013recent}).   

In previous work~\cite{yetkin.etal.oceans2016}, we show a rigorous way of computing the optimal search paths by deriving a decision-theoretic cost function that is associated with the accuracy of our estimate of the number of targets in the search environment. The work herein extends our search theory results reported in~\cite{yetkin.etal.oceans2016} by proposing a new cost function and a fast, near-optimal, planning algorithm to allow scalability of the search problem to larger areas.

Search theory has its roots in numerous civilian and military applications. The goal is either to maximize the probability of detecting the target or to minimize the expected time until a decision about the presence or absence of the target is made~\cite{chung2007decision}. In a realistic search problem, there are certain limitations on successfully locating the target such as imperfect sensor measurements or uncertain knowledge of the search environment and its affect on sensor performance. Noisy sensor measurements often include missed detections, \ie failing to detect a target that is present, and false alarms, \ie detection of a target that is not present. Local environmental conditions may also affect the number of false alarms and missed detections the sensor observes. Although there is an extensive literature on search theory (see e.g.,~\cite{benkoski1991survey, chung.etal.AR2011}), the issue of false alarms is seldom addressed.  Exceptions include~\cite{chung2012analysis} and \cite{kriheli2016optimal}, but  uncertainty in environmental conditions is not accounted for in these studies. We build upon prior work by accounting for false alarms, missed detections and uncertainty in the environment. 

The problem of computing search paths through an uncertain environment has been shown to be NP-Hard \cite{singh2009efficient}. Given the challenge of such a problem, related work in non-myopic planning for fielded systems has either yielded poor computational performance \cite{singh2009efficient}, limitations to paths that are pre-planned \cite{binney2012branch}, or employed sampling-based approaches to generate approximate solutions \cite{hollinger2015long}. In this paper, to address the problem complexity, we focus on abstracting the problem in such a way that the real-time performance for a non-myopic planner is tractable thereby enabling real-time performance on-board an AUV.

The remainder of this paper is organized as follows. In Section~\ref{sec:problemFormulation}, we formulate the search problem and describe our proposed cost function to minimize the risk of incorrectly estimating the number of targets. In Section~\ref{sec:onlineplanning}, we present an exact-solution method and propose an approximate-solution method for planning on-line search paths. Section~\ref{sec:bostondata} describes the data processing technique that was used for abstraction and mapping of the real data set, and Section~\ref{sec:numericalresults} provides numerical results.

\section{Problem Formulation}
\label{sec:problemFormulation}
\subsection{Preliminaries}
We are given a search grid $\Symbol{S} = \{s_1, s_2, \ldots, s_K \} \subset \mathbb{R}^2$ that consists of $K = n_r n_c$ disjoint cells where $n_r$ is the number of rows and $n_c$ is the number of columns in $\Symbol{S}$, and a distance constraint $\Symbol{H}$ of the mission. We associate with each cell random variables $X$ and $E$ that represent the number of targets and the environmental conditions in the cell, respectively. We presume $X_i$ is independent of $X_j$ and $E_i$ is independent of $E_j$ when $i \neq j$. The searcher's objective is to estimate $X_1, \ldots, X_K$. We assume that the search sensor's performance is dependent on the environment, and we assume that the environment in each cell is from a finite set of possible environments $\mathbf{e} = \{ b_1, b_2, \ldots, b_m \}$. That is, for all $s_i \in \Symbol{S}$, the environment is $e_i \in \mathbf{e}$. We presume that the actual environmental condition in each cell is not known, but that a probability distribution is known for each cell. The environment probability distribution for each cell $s_i \in \Symbol{S}$ is expressed $\Pi_i = [p_1(i), p_2(i), \ldots, p_m(i)]$ where $p_j(i) = P(E_i = b_j)$ is the probability that the environment in cell $s_i$ is $b_j$. 

When the search vehicle visits a location, it acquires the noisy observation $z \in Z$ of the number of targets and $y \in Y$ of the environmental conditions at that location. We use Bayesian update law to update our belief on the number of targets, $P(X)$, and the environmental conditions, $P(E)$, after acquiring $z$ and $y$ measurements. We model the likelihood of observing $z$ targets when $x$ is the true number of targets and $b_j \in \mathbf{e}$ is the true environment
\begin{align}
\P{z \mid x, b_j } & = \sum\limits_{k=0}^{\text{min}(x,z)} \binom {x} {k} D_j^k (1 - D_j)^{x-k} (1 - \alpha_j ) \alpha_j^{z-k}
\label{eq:searchsensormodel}
\end{align}
\noindent where $0 < \alpha_j \leq 1$ denote the probability of one or more false alarms and $0 < D_j \leq 1$ the probability of detection. Both $\alpha_j$ and $D_j$ are assumed to vary as functions of the environment type $b_j$. We refer the reader to~\cite{yetkin.etal.oceans2016} for more details on the observation model in~\eqref{eq:searchsensormodel}. 

After acquiring the measurements $z$ and $y$, we apply Bayes' rule to update our belief on the number of targets and on the environmental conditions.      
\begin{flalign}
& \P{x \mid z, b_j} \ \propto \ \P{z \mid x, b_j} \ \P{x \mid b_j} \label{eq:updatetargetcount} \\ 
& \P{b_j \mid y} \ \propto \ \P{y \mid b_j} \ \P{b_j} \label{eq:updateenvironment}
\end{flalign}
\noindent where $\P{z \mid x, b_j}$ is the observation model in~\eqref{eq:searchsensormodel}, and $\P{y \mid b_j}$ is the likelihood of observing the environment $y \in Y$ given true environment $b_j$. We assume that the probability of observing a particular environment is known before the mission starts and does not change. Insight on the sensor model for environment characterization arises from research on subsea bottom-type characterization, such as in~\cite{jaramillo2011auv}. We then compute the posterior belief on the number of targets unconditioned on the environment.
\begin{equation}
\P{x \mid z, y} = \sum\limits_{j = 1}^m \P{x \mid z, b_j} \P{b_j \mid y} 
\end{equation}     
\subsection{Value of Searching a Location}

After the vehicle visits a location, we compute an estimate, denoted $\delta(z)$, of the number of targets $x$ at that location, based on the observation $z$. When $\delta(z)$ is greater than $x$, we overestimate the number of targets, \ie we declare more than the actual number of targets are present. When $\delta(z)$ is less than $x$, we underestimate the number of targets, \ie we fail to declare some of the targets that are present. Both overestimation and underestimation may degrade the utility of the search results. Within a decision-theoretic framework, we define a linear loss function to penalize deviations from the true number of targets. Given the measured data $z$, we define the loss corresponding to the estimate $\delta(z)$ when $x$ is the true number of targets
\begin{equation}
\Ls{x, \delta(z) } = c_i \big \lvert x - \delta(z) \big \rvert \quad \text{for} \ i \in \{1,\ 2 \}
\label{eq:lossFunctionX}
\end{equation}

 
\noindent where $c_1 > 0$ and $c_2 > 0$ are relative costs of underestimating $(\delta(z) < x)$ and overestimating $(\delta(z) \geq x)$ the number of targets. For some subsea search applications, such as mine-hunting, overestimating the number of targets is preferred to underestimation. In our prior work we assess the utility of defining an estimate of the number of targets, while in this work we assess its loss since minimizing a risk value associated with incorrect estimation of the number of targets may be better suited for a practical application.

The posterior expected loss of computing the estimate $\delta(z)$ when the environment is $b_j$ is 
\begin{equation}
\E{\Ls{x,\delta(z)} \mid z, b_j } = \sum\limits_{x \in X} \P{x \mid z, b_j} \Ls{x, \delta(z)} 
\label{eq:expectedLossX} 
\end{equation}
\noindent where the expectation is taken over the parameter space $X$ with respect to the posterior distribution $P(x \mid z, b_j)$. The estimator $\delta^\star$ is called Bayes estimator when it minimizes the expected loss in~\eqref{eq:expectedLossX}. 
\begin{equation}
\delta^\star(z) = \arg\min_{\delta(z)} \ \E{\Ls{x, \delta(z)} \mid b_j} 
\label{eq:bayesEstimator}
\end{equation}  

Expected loss in~\eqref{eq:expectedLossX} is called Bayes' risk when $\delta(z)$ is the Bayes estimator in~\eqref{eq:bayesEstimator}. For notational convenience and the clarity of the presentation, we denote
\begin{align}
& \rho \ = \  \sum\limits_{j = 1}^m \P{b_j} \ \E{\Ls{x, \delta^\star} \mid b_j} \label{eq:priorRiskCond} \\
& \risk{z, b_j} \ = \  \E{\Ls{x, \delta^\star(z)} \mid z, b_j} \label{eq:futureRiskCond}
\end{align}

\noindent and call~\eqref{eq:priorRiskCond} the \emph{current risk} and~\eqref{eq:futureRiskCond} the \emph{anticipated risk} conditioned on search measurement $z$ and environment $b_j$. 

Due to the uncertainty in the environment and the noise in environment observations, anticipated risk in~\eqref{eq:futureRiskCond} can be different than actual risk that considers the true environment. We define a linear loss function that penalizes deviations from actual search performance. Let $e$ be the true environment in a cell. Given the observed number of targets $z$ and environmental measurement $y$, we define the loss of computing the estimate $d(y)$ on the environmental conditions
\begin{equation}
\W{ e,d(y) }  = \bar{c}_i \big \lvert \risk{z, e}  - \risk{z, d(y)} \big \rvert  \quad \text{for} \ i \in \{1,\ 2 \}
\label{eq:lossFunctionE}
\end{equation}
\noindent where $\risk{z,e}$ and $\risk{z,d(y)}$ are the anticipated risk conditioned on the observation $z$ when true environment is $e \in \mathbf{e}$ and when the environment estimate is $d(y)$, respectively. Similar to~\eqref{eq:lossFunctionX}, $\bar{c}_1, \bar{c}_2 > 0$ are the relative costs of underestimating $(d(y) \preceq e)$ and overestimating $(d(y) \succ e)$ the environmental conditions.    

Given the measurements $z$ and $y$, the posterior expected loss of computing the environment estimate $d(y)$ is
\begin{align}
\E{ \W{ e, d(y) } \mid z, y } = \sum\limits_{j = 1}^m \P{ b_j \mid y } \W{ b_j, d(y) }
\label{eq:expectedLoss}
\end{align}
An estimate of the risk associated with the incorrect estimation of the number of targets after acquiring the measurements $z$ and $y$ is
\begin{flalign}
& \risk{z, d^\star(y)} \ = \ \E{\Ls{x, \delta^\star(z)} \mid z, d^\star(y)} \label{eq:futureRiskEst} \\
& d^\star(y) \ = \ \arg\min_{d(y)} \E{ \W{ e, d(y) } \mid z, y } \label{eq:bayesEstimatorForEnv}
\end{flalign}

In order to assess the benefit of searching a location, we compute the expected loss in~\eqref{eq:futureRiskEst} for every possible set of observations $z \in Z$ and $y \in Y$. Let the anticipated risk for visiting a cell be defined as

\begin{equation}
\textbf{r}(Z,Y) = \sum\limits_z \sum\limits_y \P{ z,y } \E{\Ls{x, \delta^\star(z)} \mid z, d^\star(y)} \label{eq:anticipatedRisk}
\end{equation}
\noindent where
\begin{equation}
\P{ z, y } = \sum\limits_x \sum\limits_{b_j} \P{ x \mid z, b_j } \P{ y \mid b_j } \P{ x } \P{ b_j } 
\label{eq:PzyJoint}
\end{equation}

\noindent Further, we define the benefit of searching a location as the risk reduction we expect to achieve before seeing the measurements $z$ and $y$. That is, the value of searching cell $i$ is the difference between the current risk and the anticipated risk for that cell

\begin{equation}
B(i) = \rho_{i} - \textbf{r}_{i}(Z,Y) 
\label{eq:valueofsearch}
\end{equation}

\subsection{Planning Survey Routes to Maximize Expected Risk Reduction}

We cast the problem of finding the optimal \textit{search path} as a search over all feasible paths from the initial location of the search vehicle that satisfy the distance constraint of the mission. Let $\Symbol{H}$ be the distance constraint of the mission, $\Symbol{H}_{\gamma}$ be the distance required to traverse the path $\gamma$, and $B(i)$ in~\eqref{eq:valueofsearch} be the value of searching cell $i$. A path is then described by an ordering of cells to be inspected, $\gamma : \lbrack i , i+1,\ldots,i+h \rbrack$, such that the distance required to traverse the path $\Symbol{H}_\gamma$ is less than the mission constraint $\Symbol{H}$. The search problem is defined,
\begin{definition}
Given the maximum distance the vehicle can traverse, the value of searching each cell, and a set of feasible paths $\gamma \in \Gamma$, find the path $\gamma^\star$ such that
\begin{equation}
\gamma^\star = \argmax_{\gamma \in \Gamma} \big ( \sum\limits_{i \in \gamma} B(i) \big )
\end{equation}
\noindent subject to
\begin{equation}
\Symbol{H}_{\gamma} \leq \Symbol{H}
\end{equation}
\end{definition}
We do not allow multiple visits to the same cell.  However, re-planning the optimal path after acquiring new observations allows us to visit the same cell more than once if doing so would be beneficial.  A future goal of our research  is to incorporate multiple visits to the same cell within our computational framework. 

\section{On-line planning for real-time search}
\label{sec:onlineplanning}

Planning survey missions for AUVs in real environments requires both the consideration of the areas to be searched as well as the constraints imposed on the vehicle to acquire the data. In this work we have to contend with the constraint of acquiring acoustic data that must be spatially and temporally correlated. To execute the search sufficiently the vehicle is required to perform straight-line paths through the search region. Therefore, we constrain the vehicle to only visit cells within a single row until the row itself has been fully explored. Once the AUV has reached the end of a row, the sensor is disabled for a short period of time while it traverses to another row to perform another line measurement (see Fig.~\ref{fig:searchpaths}). 

Given this constraint, the planning problem becomes a search over the cells that can be measured within a finite time/distance horizon $\Symbol{H}$. While the planner could consider  the value of each cell independently, as the size of the search area increases, so does the complexity of the search.

\subsection{Exact Branch-and-Bound Algorithm}
\label{sec:exact_bnb}

We first apply an exact branch-and-bound algorithm with a best-first search method to compute the optimal search path. While this method yields the maximum expected risk reduction we can achieve, it is computationally less efficient compared to the approximate solution.   

Branch-and-bound algorithms are commonly used to solve large state-space search problems~\cite{wah1985stochastic}. They repeatedly partition the problem into subproblems, which is called \emph{branching}, and compute a lower and an upper bound on these subproblems, which is called \emph{bounding}. Subproblems that will not lead to the optimal solution are pruned. This reduces the computational complexity of the problem compared to brute-force search where all candidate solutions are enumerated prior to the search and the value of each of these candidate solutions are computed. To select the subproblems for branching in a systematic manner, we use a best-first search method where the most promising nodes, the nodes that have the largest upper bounds, are expanded first.  The search is performed until the search space is fully explored, yielding the optimal solution $\gamma^*$.

\subsubsection{Lower Bound} 

The lower bound used in the exact solution is calculated by summing the benefit of searching a location (Eq.\ref{eq:valueofsearch}) over a lawn-mover path, a path that consists of parallel linear tracks until the mission length is met. Since this path belongs to the set of feasible paths $\Gamma$, the optimal solution is guaranteed to yield an expected risk reduction greater than or equal to the value of this path. This value is calculated once at the start of the search in \Symbol{O}($K$) time where $K$ is the number of cells. When the value of expanded nodes of a candidate path exceeds this lower bound, we update the lower bound to be the value of this candidate path.   

\subsubsection{Upper Bound}

To compute the upper bound for the maximum expected risk reduction on a given path, the maximum expected risk reduction in a cell throughout the search area is multiplied by the number of cells the vehicle can visit without exceeding the mission length. That is, 

\begin{equation}
\textrm{UB} = \big( \argmax_{s_i \in S}B(i) \big )  \Symbol{H}
\end{equation}

As the search is expanded, the upper bound is re-computed by subtracting the maximum risk reduction in a cell from the value UB and adding the risk reduction associated with the newly added cell. The initial upper bound computation takes $\Symbol{O}(K)$ time while updates are done in $\Symbol{O}(1)$. While these bounds may not prune the search tree as effectively as tighter bounds, they take significantly less time to compute.

\subsection{Approximate Algorithm}
\label{sec:approximate_bnb}
A exhaustive search over each permutation of all feasible paths within the time horizon would in a worst case be $\mathcal{O}(K!)$. To address this we re-formulate the planning problem as a search over a graph where the vertices of the graph $v_{i} \in \mathcal{V}$ correspond to horizontal rows over the search area $\Symbol{S}$, and the edges $\epsilon_{ij} \in \mathcal{E}$ are the distance required to travel between rows. We can then define the search problem over this graph as follows: 

\begin{definition}
Given
\begin{itemize}
\item  An undirected symmetric graph $\mathcal{G} = (\mathcal{V},\mathcal{E})$, 
\item  A maximum tour length of $\mathcal{H}$, 
\item  A reward, $R_i$ associated with each vertex $v_i \in \mathcal{V}$, and
\item  A cost, $c_{i,j}$ associated with each edge $\epsilon_{i,j}\in \mathcal{E}$,
\end{itemize}
compute an ordered open tour $T^{*}=\{v_1,v_2,...,v_{h}\}$ that visits a subset of vertices $v \subseteq \mathcal{V}$ such that

\begin{equation}
v_i \neq v_j \quad \textrm{for} \; i,j \in \{1,...,h\},\; i \neq j, 
\end{equation}
\begin{equation}
v_1 \in \Symbol{T}^{*},
\end{equation}
\begin{equation}
\mathlarger{\sum}_{i=1}^{h-1} c_{i,i+1} \leq \Symbol{H},
\end{equation}
\begin{equation}
\mathlarger{\sum}_{i=1}^{h} R_i \quad \textrm{is maximized}.
\end{equation}

\label{def:defPlanning}
\end{definition}

This problem definition follows that of the \textit{Selective Traveling Salesman Problem} or \textit{Orienteering Problem} which is a special case of the \textit{Traveling Salesman Problem} and has been heavily studied in the literature (see \cite{feillet2005traveling,vansteenwegen2011orienteering} for a comprehensive survey). 
In this work the reward is a function of the sum of the expected risk reduction across the entire row a survey would take place in. Therefore, the reward for a given node is defined as:

\begin{equation}
R_i = \mathlarger{\sum}_{j=(i-1) n_c + 1}^{i n_c} B(j) \quad \quad i \leq n_r
\end{equation} 

\noindent where $n_r$ and $n_c$ are the number of rows and the number of columns in the search grid $\Symbol{S}$.

Following the upper bound on the optimal solution defined in \cite{laporte1990selective} we make a slight modification such that the tour is not constrained to ending at specific node. While the 0-1 knapsack problem is NP-Hard, only a valid upper bound is required in order to bound the Orienteering Problem. Following the approach in \cite{martello1987algorithms}, upper bounds are computed by searching for a terminal vertex in a sorted array of weighted vertices. That is, given a sorted array of non-increasing values of the reward per unit weight: 
\begin{equation}
R_j/w_j \geq R_{j+1}/w_{j+1} \quad \textrm{for} \quad j=1,\ldots,n_r-1
\label{eqn:KP01}
\end{equation}

\noindent where $w_j = \argmin_{i \neq j}(c_{i,j})$ for $j = 1, \ldots, n$. The upper bound on the reward can be computed by finding the maximum reward that can be obtained without exceeding the budget $\Symbol{H}$. Since this is a summation over the sorted array the upper bound has a complexity of $\mathcal{O}(n_r)$. Further detail on how to tighten the bounds can be found in \cite{martello1987algorithms}.

Using these upper bounds, solutions for the planning problem are found by implementing a depth-first branch-and-bound (DFBnB) solver \cite{zhang2000depth} where each node in the search tree corresponds to performing different survey lines. The branching of the tree is ordered such that nodes associated with higher reward per weight (Eq.~\ref{eqn:KP01}) are evaluated first. To prune the search based off of the maximum possible reward at each node, the upper bound previously described is employed \cite{laporte1990selective}. This bound can be re-computed during each iteration in $\mathcal{O}(n_r)$. Furthermore, each time a new tour is discovered that yields a higher reward, the ordering of the tour is optimized via 2-opt \cite{croes1958method} in order to ensure the lowest cost tour for the highest reward is being produced. Instead of performing the search exhaustively, after a large number of iterations, $\beta$, the search is terminated.  

\section{Mine Countermeasures Performance Mapping}
\label{sec:bostondata}

\begin{figure}[!t]
    \centering
  \begin{subfigure}[]{0.85\columnwidth}
       \includegraphics[width=1\columnwidth]{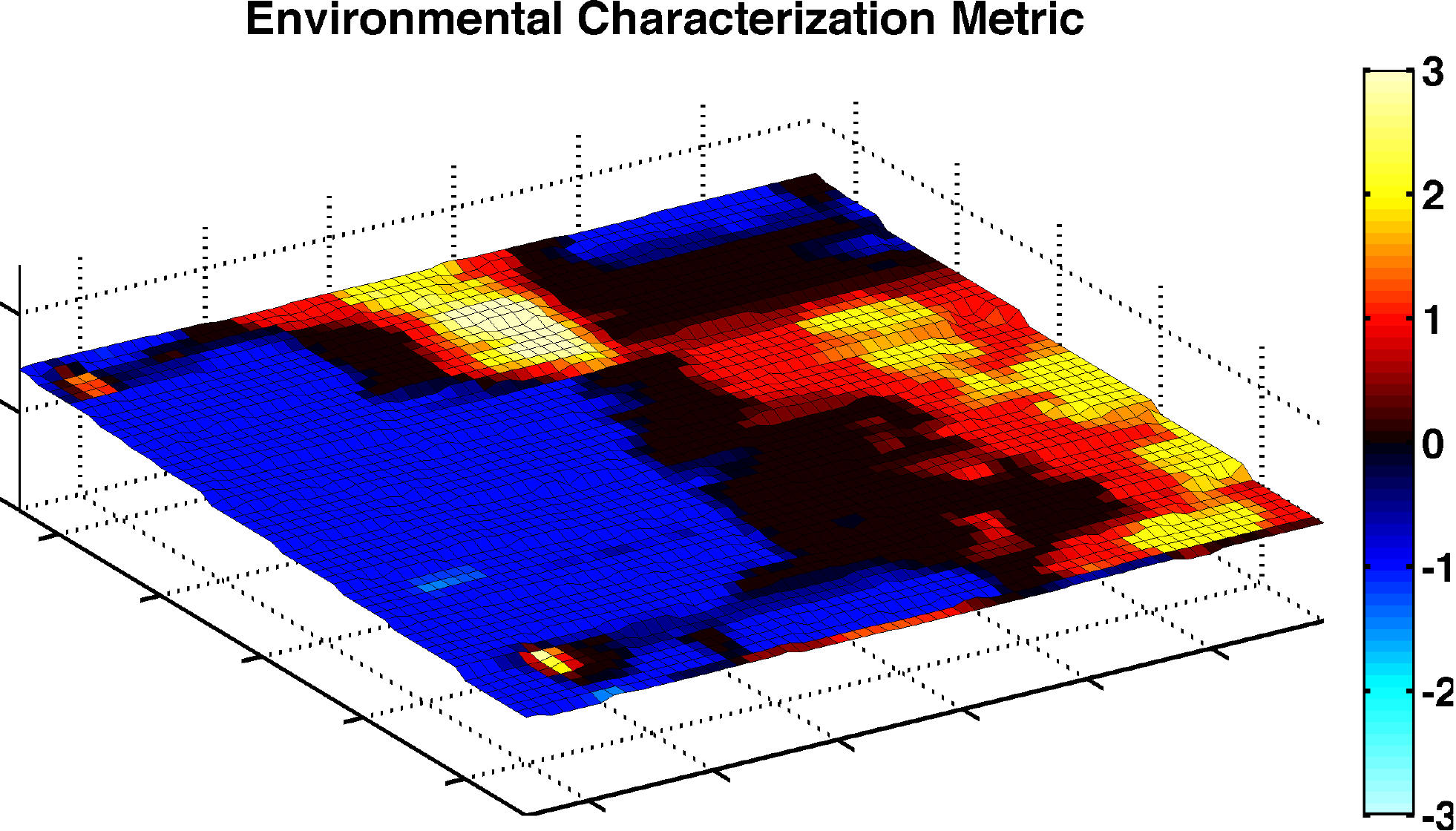}
       \subcaption{} 
  \end{subfigure}  
  \begin{subfigure}[]{0.85\columnwidth}
        \includegraphics[width=1\columnwidth]{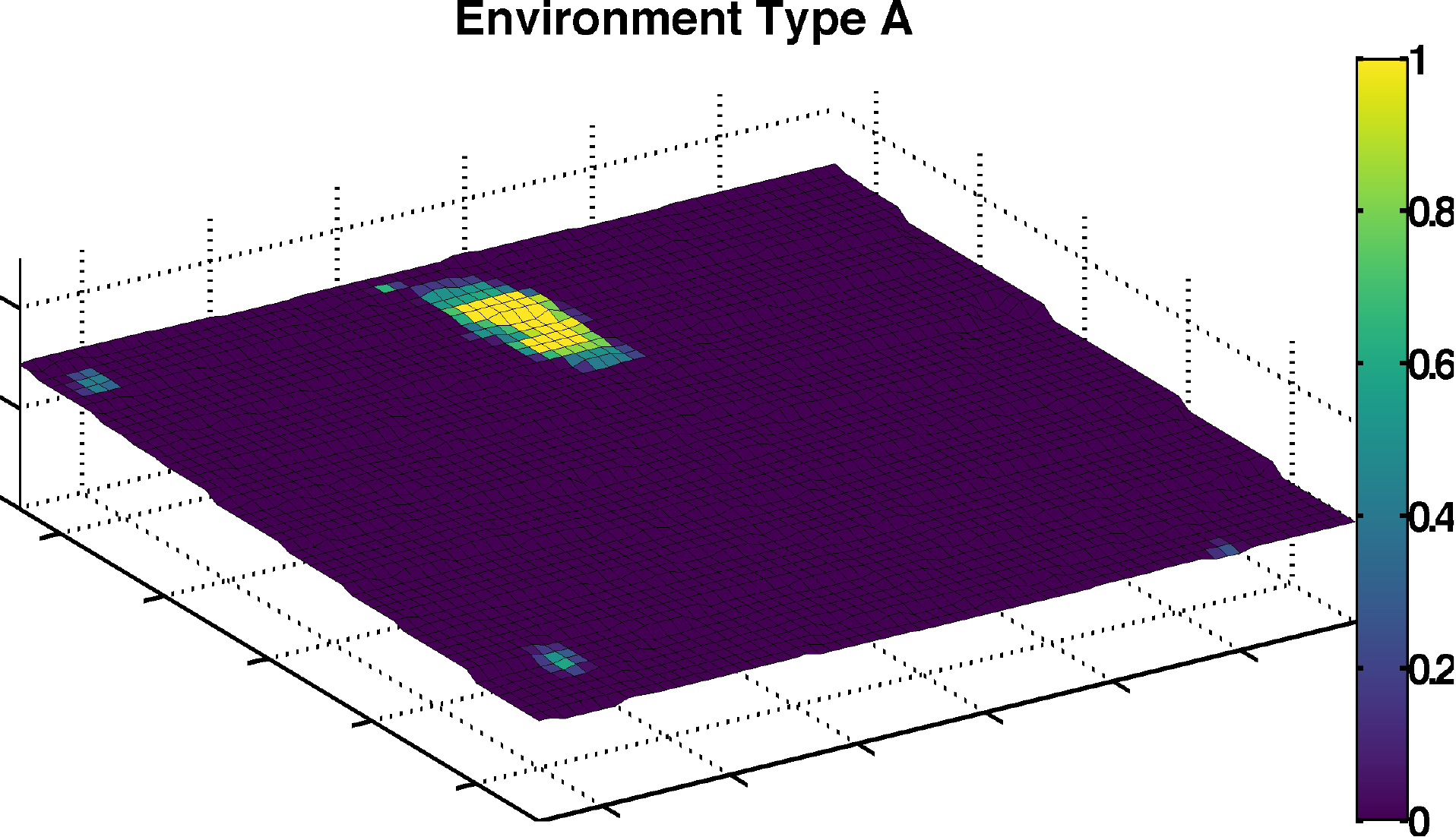}
    \subcaption{} 
  \end{subfigure}  
  \begin{subfigure}[]{0.85\columnwidth}
        \includegraphics[width=1\columnwidth]{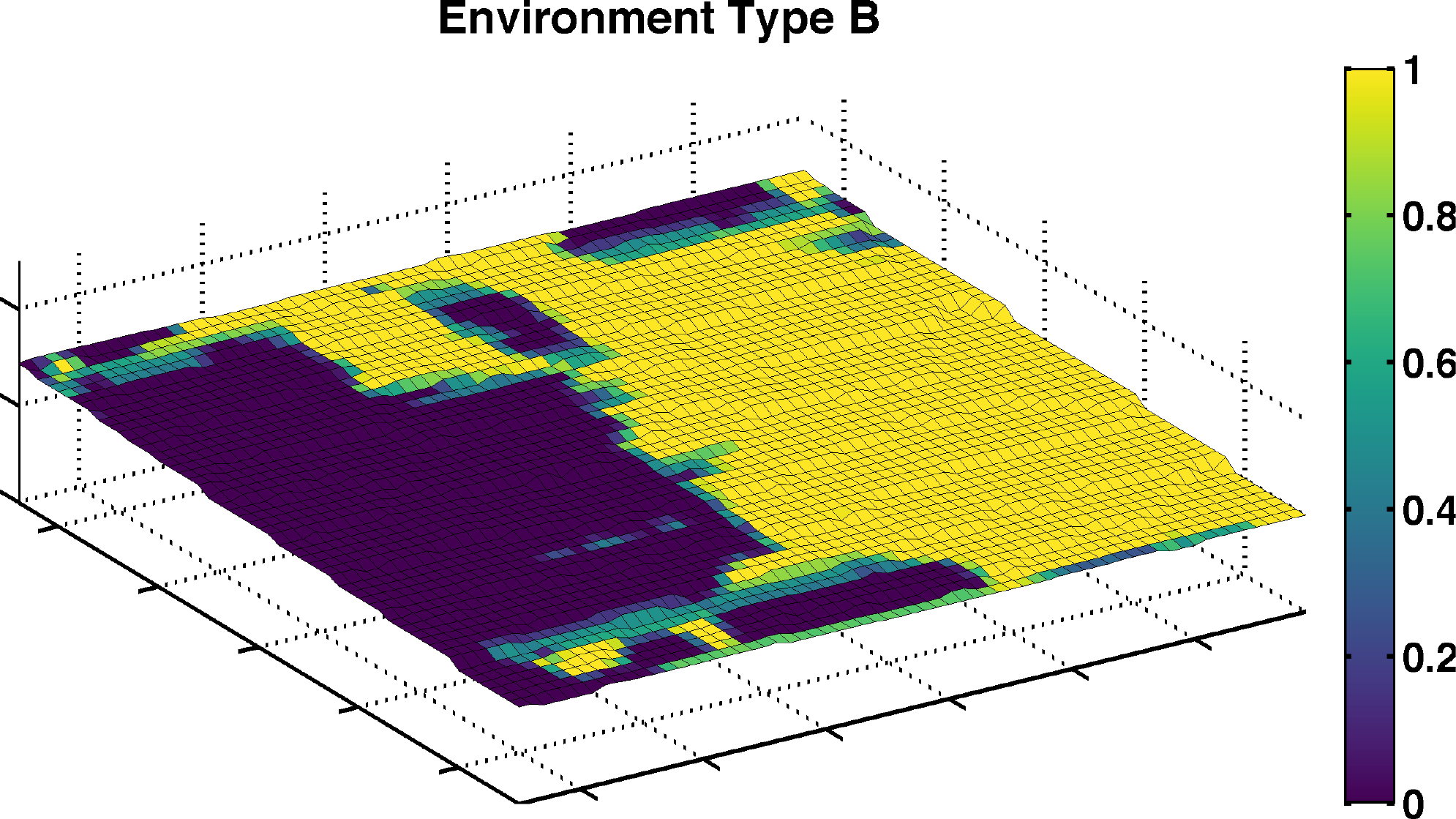}
    \subcaption{} 
  \end{subfigure}  
  \begin{subfigure}[]{0.85\columnwidth}
   \includegraphics[width=1\columnwidth]{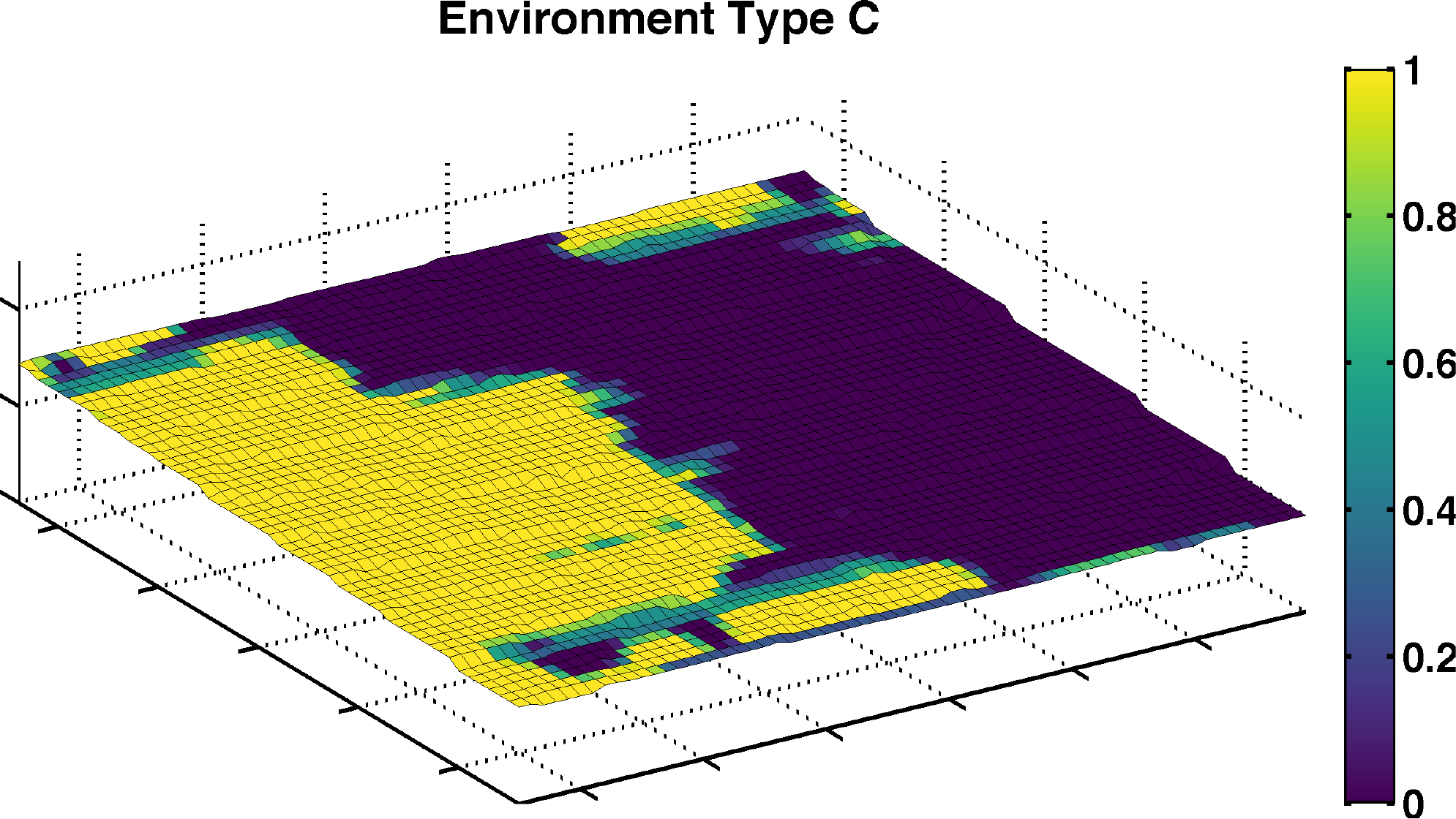}
     \subcaption{} 
  \end{subfigure}   
  \caption{ Abstraction and mapping of acoustic data collected with an autonomous underwater vehicle (AUV) to quantify mine countermeasures (MCM) performance.(a) shows the environmental characterization map. (b-d) shows the prior probability of the respective class belonging to each cell}
  \vspace{-1.5em}
      \label{fig:map}	
\end{figure}

While there are many environmental factors affecting performance, we have developed an abstraction of the acoustic data collected with the sonar based on some of these parameters. We illustrate a representative example in Fig.~\ref{fig:map}-a., and have developed a notional mapping between this abstraction and relevant mine countermeasures (MCM) performance metrics. The mapping is employed to define segmented regions of the ocean floor within which the expected mine-hunting performance may be estimated in real-time as a vehicle executes its mission.  In the following analysis, the mapping is defined by three segments characterizing the environment as difficult, moderate, or easy for MCM, as in Fig. \ref{fig:map}-b to \ref{fig:map}-d.  In this section, we describe the processing approach through which the abstraction and mappings are derived and apply the technique to data collected at-sea with a representative side-look bottom-mapping active sonar in an exercise conducted in the approaches to Boston Harbor during 2015.


\indent At the beginning of the exercise, an AUV was programmed to execute linear tracks comprising a “lawnmower” search mission over an area of the seafloor, $M \times N$ non-dimensional (grid) units in area.  As the vehicle progressed through each line, the active sonar aboard the AUV transmitted acoustic energy primarily in the broadside direction, orthogonal to the flight-path of the vehicle, and collected acoustic returns on a planar receive array. The element-level active acoustic data are first pre-processed through farfield beamforming \cite{maranda1989efficient} and matched filtering \cite{ainslie2010principles}, using techniques employed ubiquitously in modern day sonar systems.  In the analysis considered here, only the broadside monostatic return angle is retained, although it is straightforward to extend these techniques to other steering angles as well.  For each line-run, the pre-processing results in a two-dimensional grid of (unipolar) acoustic return magnitudes, $\eta_{mn}$, as a function of down-range position index, $m$, and cross-range position index, $n$, with respect to the flight path of the sonar; in the absence of background ambient noise, these data are proportional to the backscattered acoustic energy of patches on the seafloor under observation and will be referred to as such in the remainder of this section.
\newline
\indent For each line run, the beamformed and matched filtered acoustic data are abstracted as follows.  First, a two-dimensional median filter is employed to down-sample the high resolution acoustic data to length scales relevant to performance estimation.  All spatial scales in this manuscript are described in non-dimensional units as the planning algorithms are generally applicable regardless of length-scale.  After applying the median filter, for each down-range bin in the grid we compute the mean target strength over all cross-range bins over the course of an entire line run.

This results in an estimate of the acoustic energy as a function of down-range position.  Generally, this quantity was observed to be relatively stationary across all lines collected.  After subtracting the background mean at each cross-range bin, $\hat{\eta}_{mn} = \eta_{mn} - \mu_m$, we normalize by the standard deviation as a function of cross-range position index, $\Phi_{mn} = \hat{\eta}/\hat{\sigma}_{mn}$. This whitening procedure results in an environmental performance map, $\Phi_{mn}$.  A representative example for data collected during a Boston 2015 exercise is presented in Fig. \ref{fig:map}-a for all lines prosecuted; we have fused the grids in regions where the lines overlap.  The color-scale of this figure represents the number of standard deviations the background environment scattering strength deviates from the mean.  We posit that such maps are correlated with relevant MCM performance metrics, the details of which are beyond the scope of the present investigation.  We have segmented the environmental map into regions defined as difficult (Fig. \ref{fig:map}-b.), moderate (Fig. \ref{fig:map}-c.), and easy (Fig. \ref{fig:map}-d.) for MCM based upon the deviations of the background of $ \Phi_{mn} > 2$, $ \Phi_{mn} < -.5$, and $ -.5 \leq \Phi_{mn} \leq 2$ respectively.  These maps are employed in planning and performance optimization described in the remaining sections of this manuscript.  We note that the data processing described in this section has been specifically configured with the goal of real-time implementation on-board an AUV.
\section{NUMERICAL RESULTS}
\label{sec:numericalresults}
\begin{figure}[b]
\begin{minipage}{\textwidth}
\begin{minipage}[b]{.25\textwidth}
\includegraphics[width=0.8\textwidth,keepaspectratio,clip]{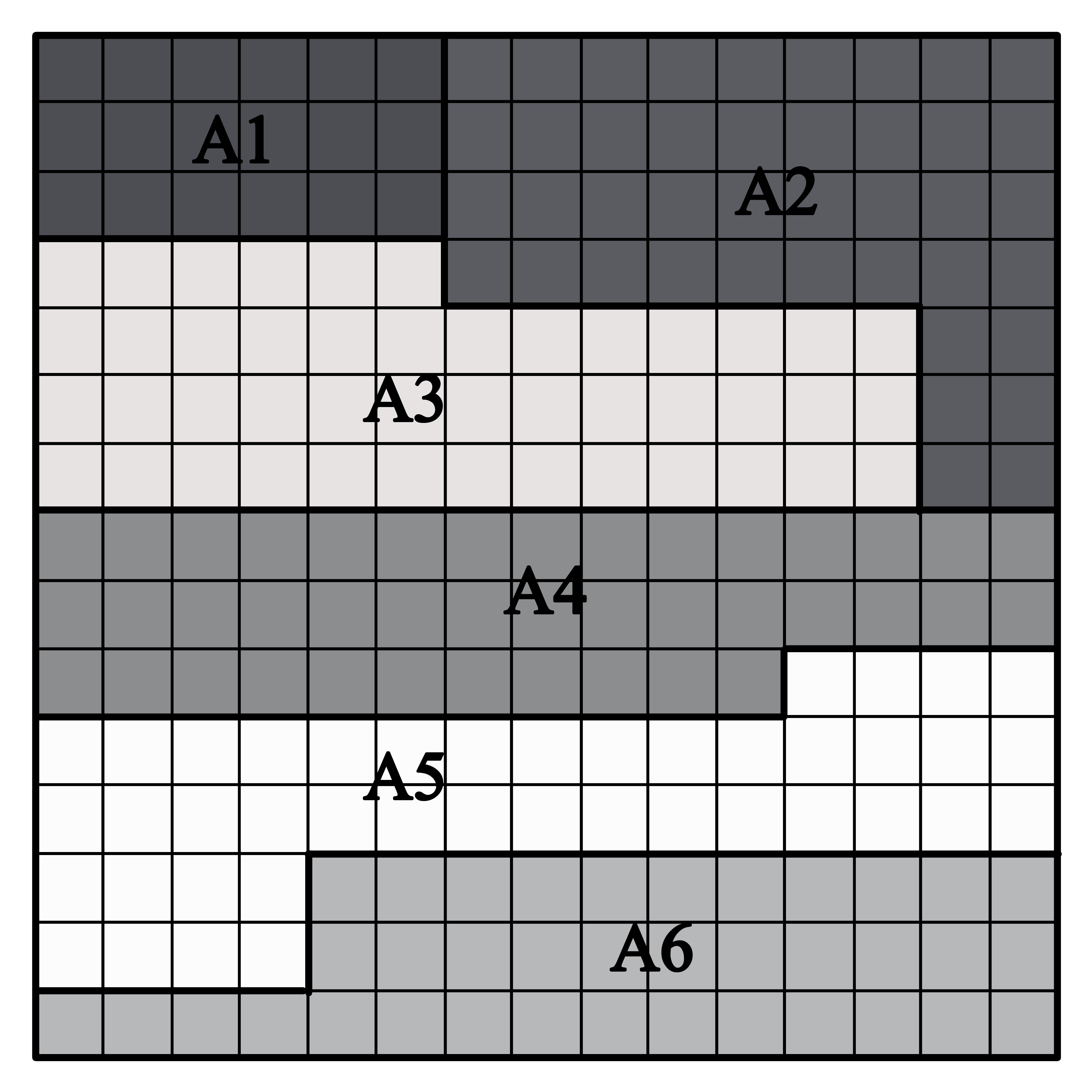}
\end{minipage}
\begin{minipage}[b]{.25\textwidth}
    \begin{tabular}{l@{\hskip 0.1cm} c @{\hskip 0.1cm}l } \vspace*{-3.4cm}
\mbox{} & \mbox{} \\  
A1 & $\rightarrow$ & $[0.95, 0.05, 0.00]$\\
A2 & $\rightarrow$ & $[0.85, 0.10, 0.05]$\\
A3 & $\rightarrow$ & $[0.10, 0.10, 0.80]$\\
A4 & $\rightarrow$ & $[0.30, 0.40, 0.30]$\\
A5 & $\rightarrow$ & $[0.00, 0.10, 0.90]$\\ 
A6 & $\rightarrow$ & $[0.10, 0.65, 0.25]$\\ 
\label{tab:distributions}
\end{tabular} 
\end{minipage}
\end{minipage}
\caption{ Search area and cell-wise environment distributions}
\label{fig:searchArea}
\end{figure}

In this section we numerically investigate the performance of the exact algorithm proposed in Section~\ref{sec:exact_bnb} and the approximate algorithm proposed in Section~\ref{sec:approximate_bnb}. First, the two approaches are compared in terms of overall search performance (i.e., achieved risk reduction) and computation time, over a small, notional, search area (Fig.~\ref{fig:searchArea}). Then, the performance of the approximate algorithm is demonstrated (in simulation) over a larger, more realistic environment (Fig.~\ref{fig:map}).

\vspace*{0.2cm}

\begin{figure*}[t]

  \captionsetup[subfigure]{aboveskip=-2pt,belowskip=0pt}
  \begin{subfigure}[t]{0.5\columnwidth}
       \includegraphics[width=1.1\columnwidth, left]{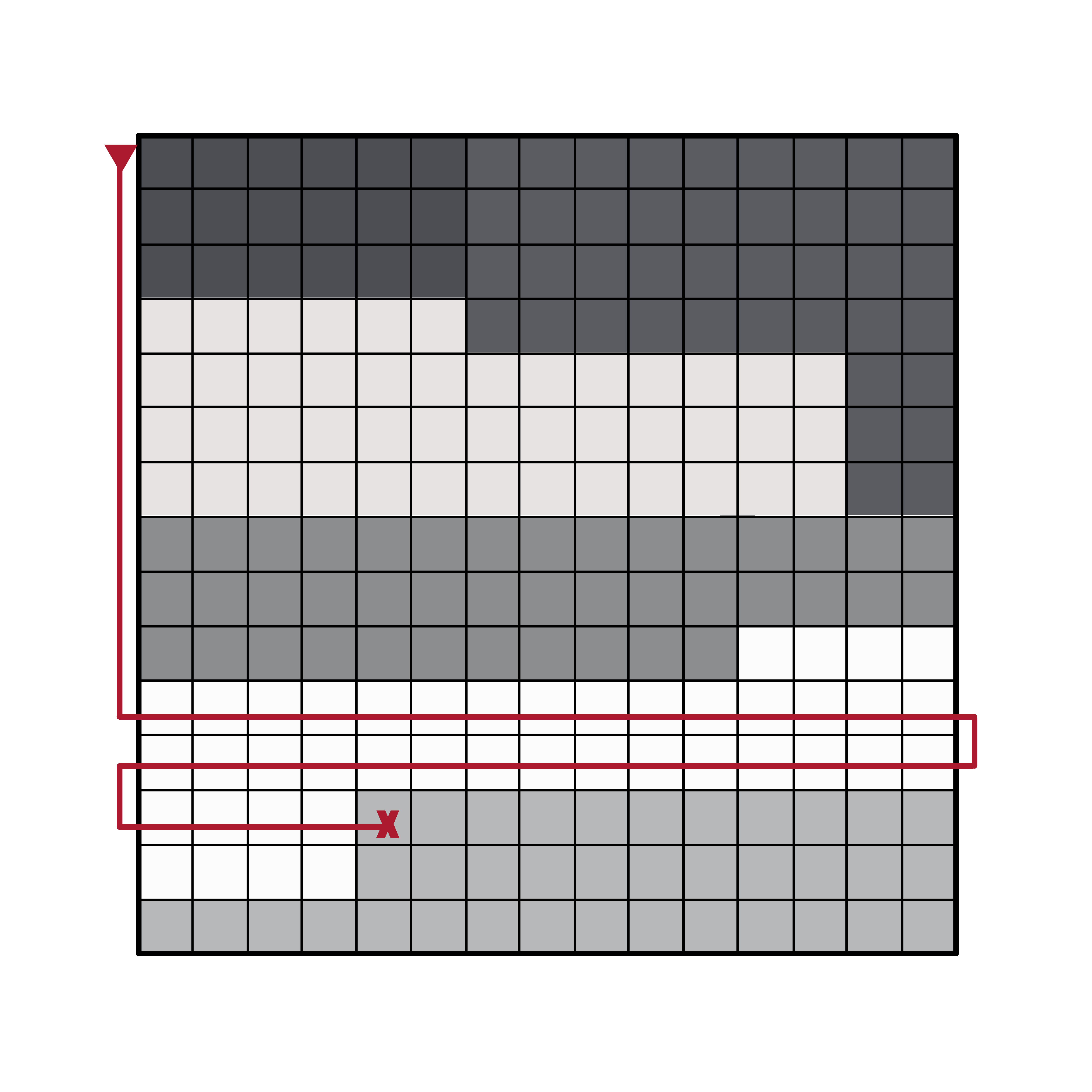}
       \caption{$\bar{B}_\text{avg}$ = 12\%}
       \label{fig:harun50}
  \end{subfigure}  
  \begin{subfigure}[t]{0.5\columnwidth}
        \includegraphics[width=1.1\columnwidth, center]{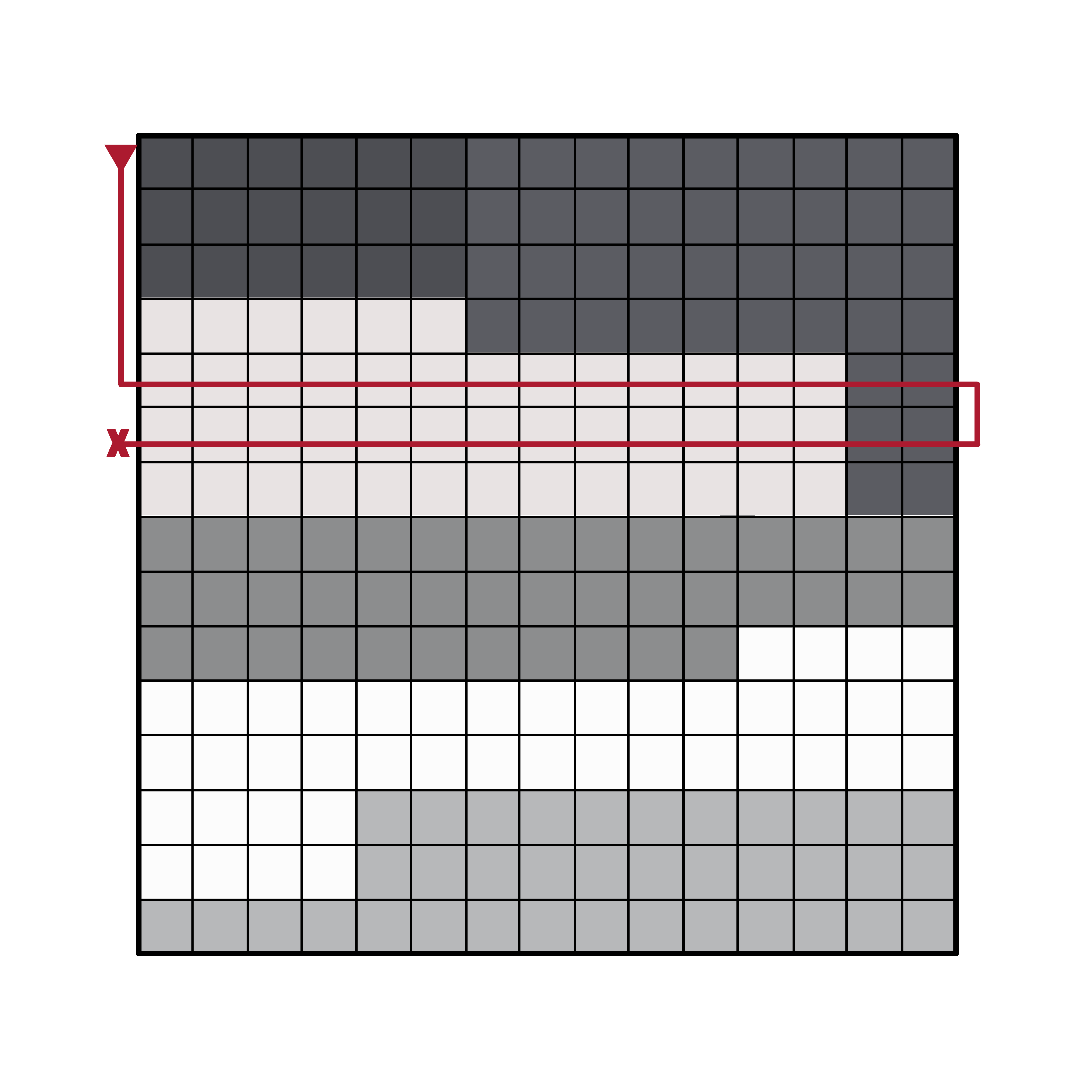}
        \caption{$\bar{B}_\text{avg}$ = 7\%}
    \label{fig:jim50}
  \end{subfigure}  
  \begin{subfigure}[t]{0.5\columnwidth}
        \includegraphics[width=1.1\columnwidth,center]{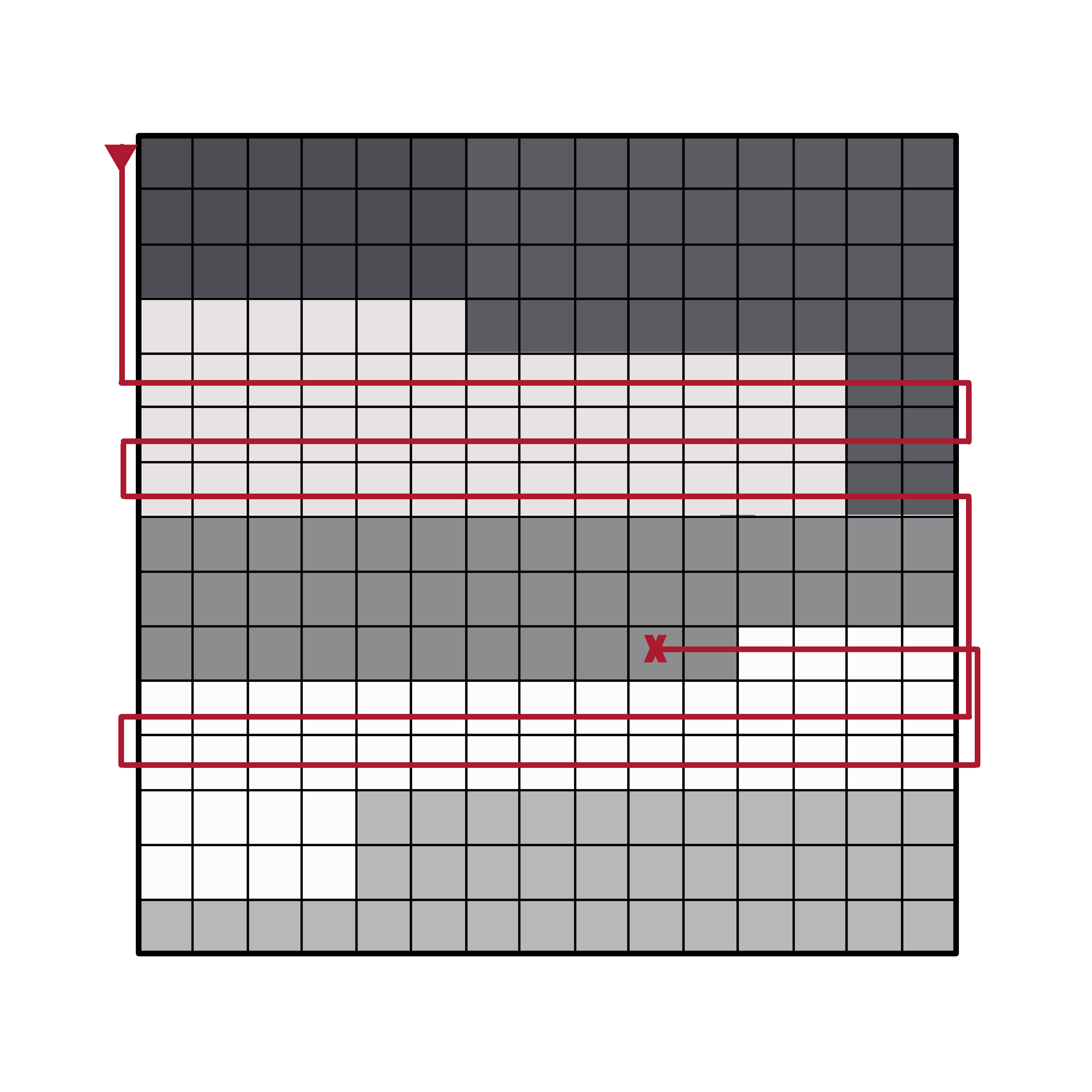}
        \caption{$\bar{B}_\text{avg}$ = 23\%}
    \label{fig:harun100} 
  \end{subfigure}  
  \begin{subfigure}[t]{0.5\columnwidth}
        \includegraphics[width=1.1\columnwidth,right]{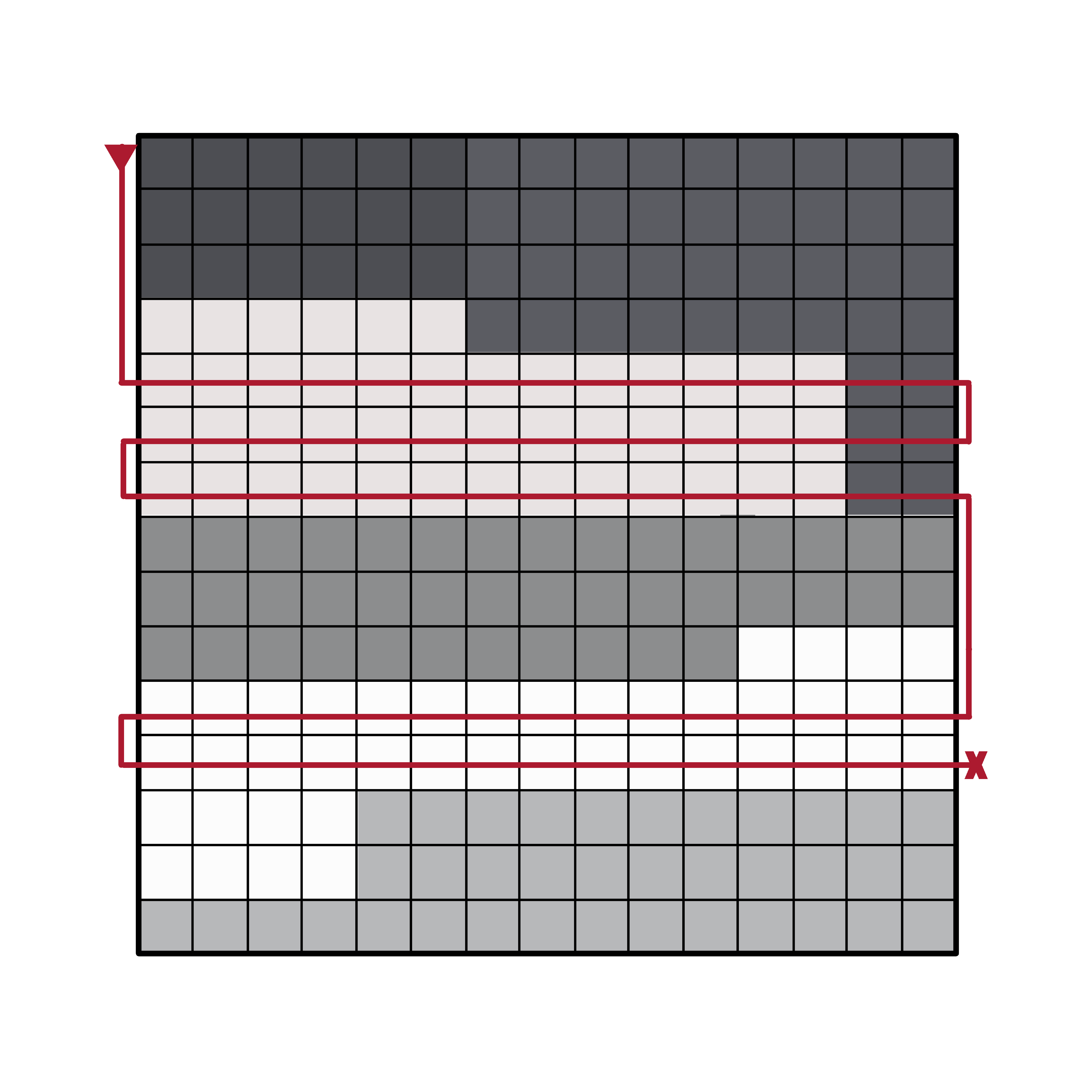}
        \caption{$\bar{B}_\text{avg}$ = 21\%}
     \label{fig:jim100} 
  \end{subfigure}   
  \caption{ Search paths for a) exact-solution method b) approximate solution method when mission length is 50 and c) exact-solution method d) approximate solution method when mission length is 100. For each path, corresponding averaged normalized risk reduction (see Fig.~\ref{fig:avgPerformance}) is also shown}
  \vspace{-1.5mm}
      \label{fig:searchpaths}	
\end{figure*}
In this study we assume that there are three candidate environments $\mathbf{e} = \{b_1, b_2, b_3 \}$. The probability of detection, $D$, and the probability of at least one false alarm, $\alpha$, for each environment is $D = 0.65$ and $\alpha = 0.4$, respectively, for environment $b_1$. Similarly, $D = 0.8$ and $\alpha = 0.3$ for environment $b_2$, and $D = 0.95$ and $\alpha = 0.05$ for environment $b_3$. Note that the information about the number of targets revealed after visiting a cell increases with increasing probability of detection and decreases with increasing probability of false alarm. Thus, environment $b_1$ is the least informative and environment $b_3$ is the most informative. We consider that the sensor model for environment characterization is 
\begin{equation}
a_{ij} = \P{E = b_j \mid Y = b_j} \ \  \text{for all } \ i,j \in \{1, 2, 3 \} 
\end{equation}
\noindent where $a_{ii}$ is the probability of observing the true environment when the environment is $b_i$. We use the sensor model with $a_{11} = 0.82$, $a_{22} = 0.84$, $a_{33} = 0.88$ and $a_{ij} = a_{ik}$ for $j,k \neq i$. 

\subsection{Comparisons of Exact and Approximate  Solutions}
\label{sec:compare}
The Monte Carlo based comparison between the exact and approximate algorithms was conducted over a 15-by-15 cell search area with known cell-wise environment distributions as shown in Fig.~\ref{fig:searchArea}. The search area is divided into 6 parts: A1 through A6. For each part, the corresponding probability distribution $\Pi = [p_1, p_2, p_3]$ is given, where $p_j$ is the probability that the environment is $b_j$. For example, for the cells labeled as A1, there is a 0.95 probability that the environment is $b_1$ and a 0.05 probability that the environment is $b_2$. On the other hand, for the cells labeled as A5, there is a 0.9 probability that the environment is $b_3$ and a 0.1 probability that the environment is $b_2$. For each of the probability distributions in Fig.~\ref{fig:searchArea}, the value of searching the corresponding cell (Eq.~\ref{eq:valueofsearch}) is different. Lighter colored cells (\eg cells contained in A5 and A3) are expected to yield greater risk reduction since there is higher probability that the environment in these cells is environment type $b_3$. Thus, these cells are preferred for search\footnote{We note that the search area in Fig.~\ref{fig:searchArea}. is selected to illustrate that the performance when using the approximate method can be suboptimal. Choosing a random search area is likely to reduce the difference between the performance of the exact method and the performance of the approximate  method. For example, rotating this search area $90\degree$ would result in less deviation from the optimal performance}     

Based on this environmental distribution a total of 500 synthetic environments (or ``scenes'') were generated for a particular mission length. For each scene, a corresponding set of deterministic measurements were generated based on the prior belief on the number of targets and the environmental conditions. Multiple measurements were simulated at each cell location to account for cases in which the same cell is visited multiple times. The mission length (in terms of number of cells) was varied in increments of 10 from 50 to 110. The scene, deterministic measurements, and a specified mission length were then passed to both solvers (corresponding to the exact and approximate algorithms) consecutively. At each iteration the respective solvers recorded the sequences of planned paths, the computation time (for the re-planning step), and the search performance. The algorithms were implemented in C++ using the Armadillo linear algebra library~\cite{armadillo} on a 64-bit Ubuntu 16.04 operation system.  All computations described in this paper were executed on an Intel Xeon E5-1650V3 processor with a processor base frequency of 3.5 GHz and 32 GB of RAM.

\begin{figure}[!b]
    \centering
  \begin{subfigure}[]{0.75\columnwidth}
       \includegraphics[width=1\columnwidth]{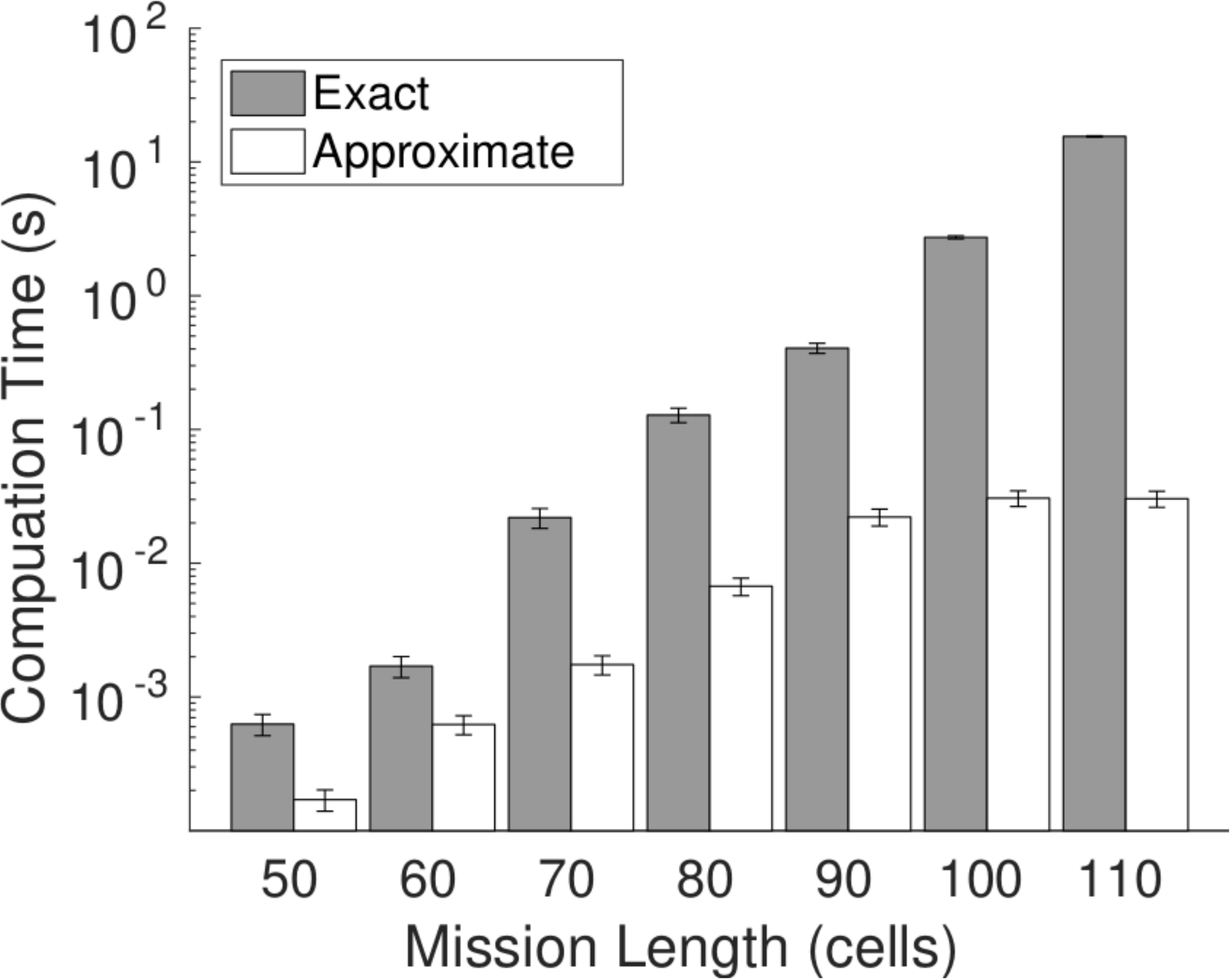}
  \end{subfigure}  
  \begin{subfigure}[]{0.75\columnwidth}
       \includegraphics[width=1\columnwidth]{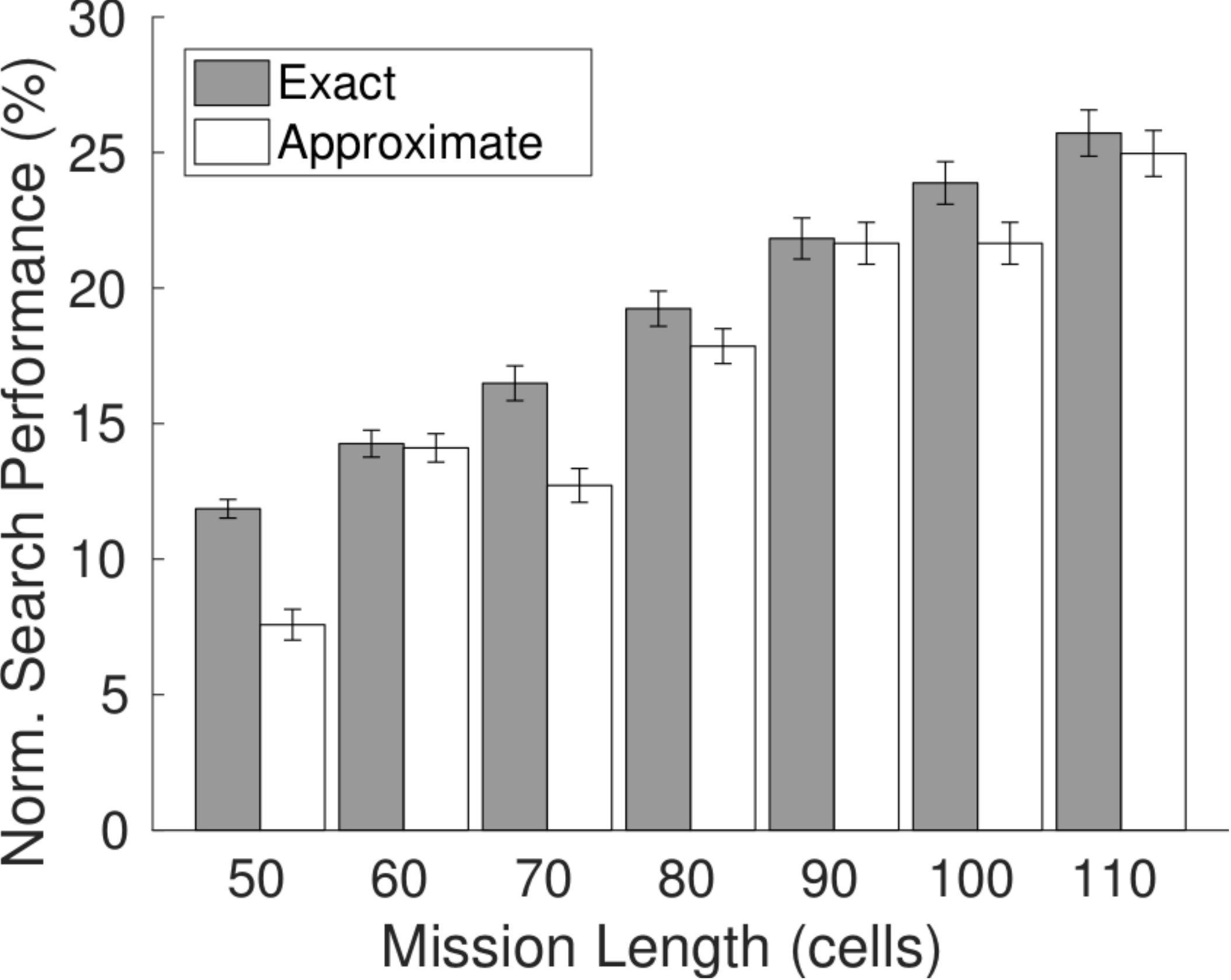}
  \end{subfigure} 
  \vspace*{3mm}
  \caption{ Comparison of average computation time (top) and average normalized search performance (bottom) between exact and approximate algorithms based on Monte Carlo simulation. Error bars indicate one standard deviation.}
\vspace{-1.5mm}
\label{fig:avgPerformance}
\end{figure}

    

The results of the Monte Carlo simulation are shown in Fig.~\ref{fig:avgPerformance}. As expected, the search performance and computation time increased with mission length for both algorithms. We found that, in general, the approximate algorithm had comparable search performance to the exact algorithm with a substantially lower computation time. The difference in computation time is particularly pronounced as the mission length increases (e.g., for a mission length of 100 cells the difference is about two orders of magnitude). The approximate algorithm is further evaluated using a more realistic environment as described in the following section. We note that in Fig.~\ref{fig:avgPerformance}, achieved risk reduction after a mission is normalized with the risk prior to acquiring any measurement. That is, let $\gamma_j$ be the path the vehicle traverses at the $j$th run, the averaged normalized achieved risk reduction $\bar{B}_\text{avg}$ is

\begin{equation}
\bar{B}_\text{avg} =  \frac{1}{\lvert \text{MC} \rvert}  \sum\limits_{j = 1}^{\lvert \text{MC} \rvert} \Big ( \frac{\sum_{i \in \gamma_j} B(i)}{\sum_{i \in \Symbol{S}} \rho_i} \Big )
\label{eq:avgNormSearchPerformance}
\end{equation}

\noindent where $\lvert \text{MC} \rvert$ is the number of Monte Carlo iterations. Thus, Fig.~\ref{fig:avgPerformance} gives the percentage of the achieved risk reduction over a long run (\eg, on average, both approaches yield above 20\% risk reduction throughout the search area).

Fig.~\ref{fig:searchpaths} shows the search paths for both algorithms for a mission length of 50 (Fig.~\ref{fig:searchpaths}a-b) and for a mission length of 100 (Fig.~\ref{fig:searchpaths}c-d). For each path, the averaged normalized achieved risk reduction (Eq.~\ref{eq:avgNormSearchPerformance}) associated with the corresponding mission length is also presented. The path traversed by the vehicle is represented by the red line. The results show that while the exact method yields the optimal search path, the approximate method results in a near-optimal search path. For both methods, the vehicle skips the parts of the search area where expected risk reduction is relatively small. We note that since the approximate method considers each row as a vertex, it does not start a row if it will not finish it. Thus, when using the approximate method, search may stop before the mission length is met.

\begin{figure*}[]
\centering
  \includegraphics[width=0.65\linewidth]{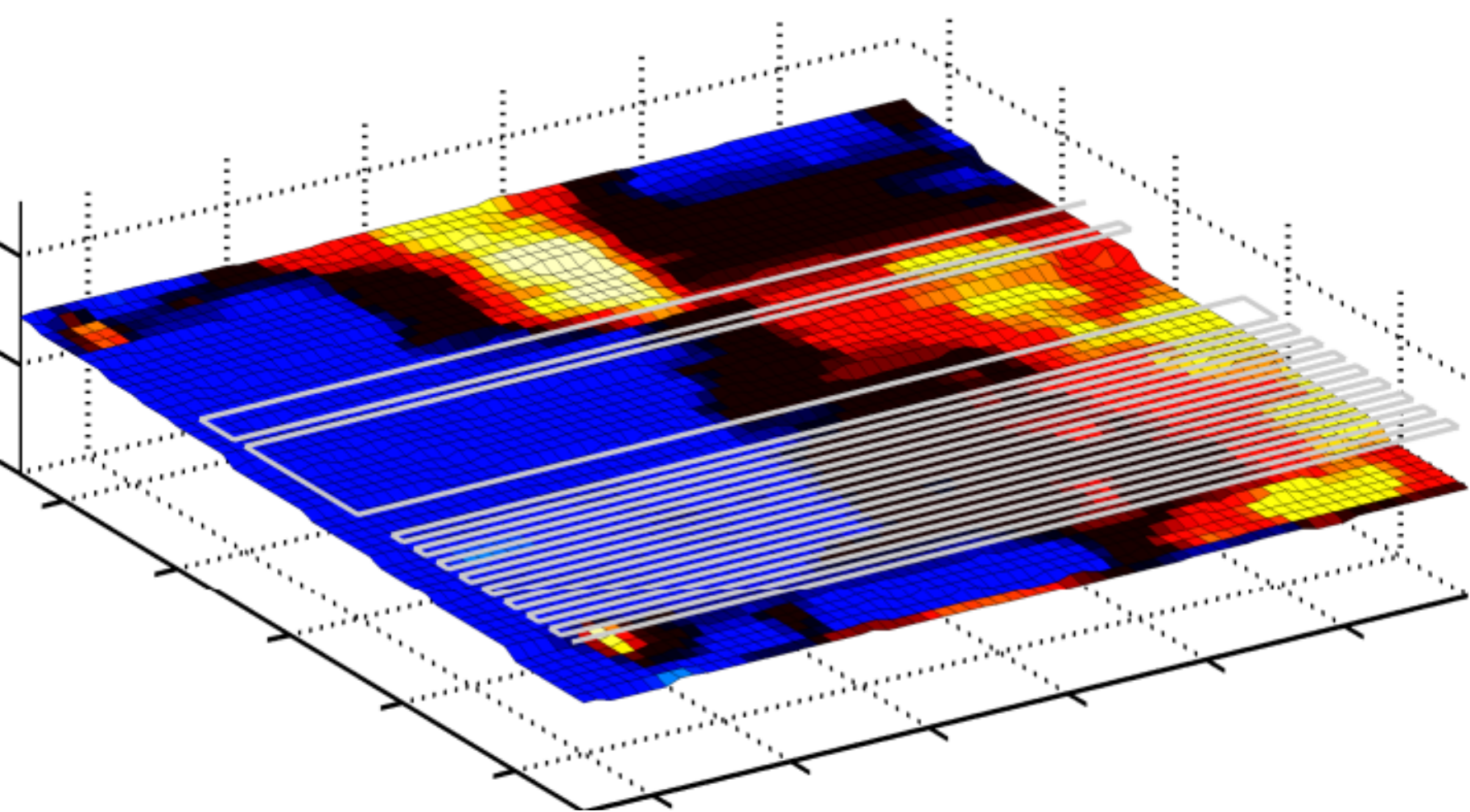} \vspace*{2mm}
  \caption{Representative planning solution for the approximate method over the large-scale environmental map ($51 \times 65$ cells) which yields a nominal search performance of $23\%$}
  \label{fig:real_solution}
  \vspace{-2mm}
\end{figure*}

\subsection{On-line Planning}
Using the environmental characterization map from the Boston Harbor discussed in Section \ref{sec:bostondata}, the approximate algorithm was used to generate search paths over the search area in a simulated re-planning framework. This re-planning framework has four steps: (i) Plan search path over search area to maximize the reduction in risk, (ii) survey first \textit{leg} of the mission, (iii) update cells with collected measurements, and (iv) return to (i). The process is repeated until the planner is no longer able to generate search paths due to insufficient planning length (\textit{the planning budget is updated each planning cycle to maintain the overall mission length}). The sensor model used and method of simulating measurements follows that of Section \ref{sec:compare}. Additionally we set $\beta = 1,000,000$ in order to provide a feasible branching factor that would achieve runtime performance on-board an AUV.  

The search area is consisted of 51 rows and 65 columns; a total number of 3315 cells. Each cell is 25m by 25m. Figure \ref{fig:real_solution} shows the final executed trajectory for a mission length of 1500 cells. The mission consisted of measuring 22 rows while re-planing after measuring each row. It can be seen that the plan covers a large portion of the \textit{easy} environment type at the start location (bottom-left corner) and eventually skips a large portion of the map to inspect another large swath of easy environment type. \footnote{We note that since the search mission is performed within limited time, searching a portion of the map with the \emph{difficult} environment type would result in reducing less risk compared to skipping that portion and performing search on easy locations} The worst-case computation time (initial plan) took approximately $0.6$s, and the final nominal search performance was $23\%$.


\section{CONCLUSIONS}

In this paper we address the problem of finding an unknown number of stationary targets within a bounded search area within bounded time. We develop an exact and an approximate algorithm for computing time-constrained search paths over the search area. Comparing both the exact and approximate algorithms, we show a near-optimal reduction in risk is achieved when using the approximate algorithm which has a runtime suitable for use on-board a fielded AUV. 

The algorithms in this work open up additional research avenues which include improving the computational complexity or solution quality of the approximate algorithm. This can be done by employing tighter lower and upper bounds on estimated mission performance as well as implementing probabilistic sampling strategies for searching the set of feasible paths. Additional performance increases can also be obtained by considering the remaining search budget in cases where the whole-row approximation does not exhaust the overall budget. Furthermore, considering multiple visits to single cells or correlation between cells during the planning process are also potential areas of investigation. 

\section{Acknowledgements}

The authors would like to thank Dr. Stefan Edelkamp for providing the branch-and-bound software that was used as the base for the implementation in this paper. 


\bibliographystyle{IEEEtran}


\bibliography{main}

\begin{thebibliography}{10}
\providecommand{\url}[1]{#1}
\csname url@samestyle\endcsname
\providecommand{\newblock}{\relax}
\providecommand{\bibinfo}[2]{#2}
\providecommand{\BIBentrySTDinterwordspacing}{\spaceskip=0pt\relax}
\providecommand{\BIBentryALTinterwordstretchfactor}{4}
\providecommand{\BIBentryALTinterwordspacing}{\spaceskip=\fontdimen2\font plus
\BIBentryALTinterwordstretchfactor\fontdimen3\font minus
  \fontdimen4\font\relax}
\providecommand{\BIBforeignlanguage}[2]{{%
\expandafter\ifx\csname l@#1\endcsname\relax
\typeout{** WARNING: IEEEtran.bst: No hyphenation pattern has been}%
\typeout{** loaded for the language `#1'. Using the pattern for}%
\typeout{** the default language instead.}%
\else
\language=\csname l@#1\endcsname
\fi
#2}}
\providecommand{\BIBdecl}{\relax}
\BIBdecl

\bibitem{yetkin.etal.oceans2015}
H.~Yetkin, C.~Lutz, and D.~J. Stilwell, ``Utility-based adaptive path planning
  for subsea search,'' in \emph{OCEANS 2015 - MTS/IEEE Washington}, Oct 2015,
  pp. 1--6.

\bibitem{yetkin.etal.oceans2016}
------, ``Acquiring environmental information yields better anticipated search
  performance,'' in \emph{OCEANS 2016 - MTS/IEEE Monterey}, Sept 2016, pp.
  1--6.

\bibitem{chapple2008automated}
P.~Chapple, ``Automated detection and classification in high-resolution sonar
  imagery for autonomous underwater vehicle operations,'' Maritime Operations
  Division Defence Science and Technology Organisation, Edinburgh, Australia,
  Tech. Rep., 2008.

\bibitem{Houston2008}
B.~H. Houston, J.~A. Bucaro, T.~Yoder, L.~Kraus, J.~Tressler, J.~Fernandez,
  T.~Montgomery, and T.~Howarth, ``Broadband low frequency sonar for
  non-imaging based identification,'' in \emph{OCEANS 2002 - MTS/IEEE Biloxi},
  Oct 2002, pp. 383--387.

\bibitem{Hayes2009}
M.~P. Hayes and P.~T. Gough, ``Synthetic aperture sonar: A review of current
  status,'' \emph{IEEE Journal of Oceanic Engineering}, vol.~34, no.~3, pp.
  207--224, July 2009.

\bibitem{ouchi2013recent}
K.~Ouchi, ``Recent trend and advance of synthetic aperture radar with selected
  topics,'' \emph{Remote Sensing}, vol.~5, no.~2, pp. 716--807, 2013.

\bibitem{chung2007decision}
T.~H. Chung and J.~W. Burdick, ``A decision-making framework for control
  strategies in probabilistic search,'' in \emph{Proc. IEEE International
  Conference on Robotics and Automation}, April 2007, pp. 4386--4393.

\bibitem{benkoski1991survey}
S.~J. Benkoski, M.~G. Monticino, and J.~R. Weisinger, ``A survey of the search
  theory literature,'' \emph{Naval Research Logistics (NRL)}, vol.~38, no.~4,
  pp. 469--494, 1991.

\bibitem{chung.etal.AR2011}
T.~H. Chung, G.~A. Hollinger, and V.~Isler, ``Search and pursuit-evasion in
  mobile robotics,'' \emph{Autonomous robots}, vol.~31, no.~4, pp. 299--316,
  2011.

\bibitem{chung2012analysis}
T.~H. Chung and J.~W. Burdick, ``Analysis of search decision making using
  probabilistic search strategies,'' \emph{IEEE Transactions on Robotics},
  vol.~28, no.~1, pp. 132--144, Feb 2012.

\bibitem{kriheli2016optimal}
B.~Kriheli, E.~Levner, and A.~Spivak, ``Optimal search for hidden targets by
  unmanned aerial vehicles under imperfect inspections,'' \emph{American
  Journal of Operations Research}, vol.~6, no.~2, p. 153, 2016.

\bibitem{singh2009efficient}
A.~Singh, A.~Krause, C.~Guestrin, and W.~J. Kaiser, ``Efficient informative
  sensing using multiple robots,'' \emph{Journal of Artificial Intelligence
  Research}, vol.~34, pp. 707--755, 2009.

\bibitem{binney2012branch}
J.~Binney and G.~S. Sukhatme, ``Branch and bound for informative path
  planning,'' in \emph{IEEE International Conference on Robotics and
  Automation}, May 2012, pp. 2147--2154.

\bibitem{hollinger2015long}
G.~A. Hollinger, ``Long-horizon robotic search and classification using
  sampling-based motion planning,'' in \emph{Proc. Robotics: Science and
  Systems}, July 2015.

\bibitem{jaramillo2011auv}
S.~Jaramillo and G.~Pawlak, ``{AUV}-based bed roughness mapping over a tropical
  reef,'' \emph{Coral Reefs}, vol.~30, no.~1, pp. 11--23, 2011.

\bibitem{wah1985stochastic}
B.~W. Wah and C.~F. Yu, ``Stochastic modeling of branch-and-bound algorithms
  with best-first search,'' \emph{IEEE Transactions on Software Engineering},
  vol. SE-11, no.~9, pp. 922--934, Sept 1985.

\bibitem{feillet2005traveling}
D.~Feillet, P.~Dejax, and M.~Gendreau, ``Traveling salesman problems with
  profits,'' \emph{Transportation Science}, vol.~39, no.~2, pp. 188--205, 2005.

\bibitem{vansteenwegen2011orienteering}
P.~Vansteenwegen, W.~Souffriau, and D.~Van~Oudheusden, ``The orienteering
  problem: A survey,'' \emph{European Journal of Operational Research}, vol.
  209, no.~1, pp. 1--10, 2011.

\bibitem{laporte1990selective}
G.~Laporte and S.~Martello, ``The selective traveling salesman problem,''
  \emph{Discrete Applied Mathematics}, vol.~26, no. 2-3, pp. 193--207, 1990.

\bibitem{martello1987algorithms}
S.~Martello and P.~Toth, ``Algorithms for knapsack problems,''
  \emph{North-Holland Mathematics Studies}, vol. 132, pp. 213--257, 1987.

\bibitem{zhang2000depth}
W.~Zhang, ``Depth-first branch-and-bound versus local search: A case study,''
  in \emph{National Conference on Artificial Intelligence}.\hskip 1em plus
  0.5em minus 0.4em\relax Austin, TX, USA: AAAI, 2000, pp. 930--935.

\bibitem{croes1958method}
G.~A. Croes, ``A method for solving traveling-salesman problems,''
  \emph{Operations research}, vol.~6, no.~6, pp. 791--812, 1958.

\bibitem{maranda1989efficient}
B.~Maranda, ``Efficient digital beamforming in the frequency domain,''
  \emph{The Journal of the Acoustical Society of America}, vol.~86, no.~5, pp.
  1813--1819, 1989.

\bibitem{ainslie2010principles}
M.~A. Ainslie, \emph{Principles of Sonar Performance Modelling}.\hskip 1em plus
  0.5em minus 0.4em\relax Springer, 2010.

\bibitem{armadillo}
C.~Sanderson, ``Armadillo: An open source c++ linear algebra library for fast
  prototyping and computationally intensive experiments,'' 2010.

\end{thebibliography}

\end{document}